\documentclass[sigconf]{acmart}

\usepackage{enumitem}
\usepackage{multirow} % 用于合并表格行
\usepackage{booktabs} % 用于创建更美观的表格线
\usepackage{multirow} % 用于合并表格行
\usepackage{tabularx} % 用于创建自适应页宽的表格
\usepackage{ragged2e} % 用于更好的对齐
\usepackage{graphicx}
\usepackage{subcaption} % 推荐，用于子图
\usepackage{algorithm}
\usepackage{algorithmic}
\usepackage{amsthm}
\usepackage{enumitem}
\usepackage[table]{xcolor} % 用于单元格上色
\usepackage{amsmath}       % 用于数学符号

\AtBeginDocument{%
  }

\setcopyright{acmlicensed}
\copyrightyear{2027}
\acmYear{2027}
\acmDOI{XXXXXXX.XXXXXXX}
\acmConference[arXiv Preprint]{arXiv Preprint}{2026}{arXiv.org}

\acmISBN{978-1-4503-XXXX-X/2018/06}

\begin{document}

%%
%% The "title" command has an optional parameter,
%% allowing the author to define a "short title" to be used in page headers.
% \title{Beyond Single-Episode Optimization: Hierarchical Generative Auto-Bidding under Sliding-Window Constraints}
\title{Beyond Single-Episode Optimization: Sliding-Window Aware Generative Auto-Bidding for Long-Term Advertising Effectiveness
% \yy{Beyond Single-Episode Optimization: Future-Aware Generative Auto-Bidding via Sliding-Window Foresight}
}

%%
%% The "author" command and its associated commands are used to define
%% the authors and their affiliations.
%% Of note is the shared affiliation of the first two authors, and the
%% "authornote" and "authornotemark" commands
%% used to denote shared contribution to the research.
\author{Binglin Wu}
\email{binglin@mail.dlut.edu.cn}
\orcid{0009-0000-2295-6773}
\authornotemark[1]
\authornotemark[2]
\affiliation{%
  \institution{Dalian University of Technology}
  \city{Dalian}
  \country{China}
}

\author{Chuan Yue}
\email{yuechuan.yc@alibaba-inc.com}
\authornotemark[2]
\affiliation{%
  \institution{Alibaba International Digital
Commerce Group}
  \city{Hangzhou}
  \country{China}
}

\author{Yingyi Zhang}
\email{yingyizhang@mail.dlut.edu.cn}
\orcid{0000-0001-9062-3428}
\affiliation{%
  \institution{Dalian University of Technology \&
City University of Hong Kong}
  \city{Dalian}
  \country{China}
}

\author{Xianneng Li}
\email{xianneng@dlut.edu.cn}
\orcid{0000-0003-4130-6930}
\correspondingauthor
\affiliation{%
  \institution{Dalian University of Technology}
  \city{Dalian}
  \country{China}
}

\author{Ruyue Deng}
\email{dengruyue.dry@alibaba-inc.com}
\affiliation{%
  \institution{Alibaba International Digital
Commerce Group}
  \city{Hangzhou}
  \country{China}
}

\author{Weiru Zhang}
\email{weiru.zwr@alibaba-inc.com}
\affiliation{%
  \institution{Alibaba International Digital
Commerce Group}
  \city{Hangzhou}
  \country{China}
}

\author{Xiaoyi Zeng}
\email{yuanhan@taobao.com}
\affiliation{%
  \institution{Alibaba International Digital
Commerce Group}
  \city{Hangzhou}
  \country{China}
}

% \author{Anonymous authors}
% \affiliation{
%     \institution{Paper under double-blind review}
%     \country{}
% }

%%
%% By default, the full list of authors will be used in the page
%% headers. Often, this list is too long, and will overlap
%% other information printed in the page headers. This command allows
%% the author to define a more concise list
%% of authors' names for this purpose.
\renewcommand{\shortauthors}{Binglin Wu et al.}
% \renewcommand{\shortauthors}{Annoymous Author, et al.}

%%
%% The abstract is a short summary of the work to be presented in the
%% article.

\begin{abstract}
Auto-bidding systems optimize bids to maximize value under efficiency constraints such as Cost-Per-Action (CPA). Existing methods treat each day as an independent episode. However, many advertisers produce value so sparsely that per-day efficiency ratios become statistically unreliable, undermining advertiser retention. Platforms therefore evaluate window-level efficiency over sliding windows of $W{=}7$ days, ensuring fair evaluation and long-term advertising effectiveness. This creates cross-episode coupling: each day's bidding decisions affect up to $W$ overlapping windows, so setting daily targets requires anticipating future market conditions. We propose SWAG-Bid (Sliding-Window Aware Generative Auto-Bidding), a hierarchical framework decomposing this challenge into episode-level planning and step-level execution. The planner uses a Masked Trajectory Model to forecast markets and generate candidate plans, scored across all overlapping windows by Multi-Window Model Predictive Control Sampling (MWMS) with exponential confidence decay. The controller adjusts reliance on this guidance through a state-adaptive gate, Per-Step Gated Adaptive Layer Normalization (PSG-AdaLN), complemented by Return-to-Go and Cost-to-Go channels carrying budget and constraint information. Experiments on AuctionNet-Sparse and online A/B tests on AliExpress show that SWAG-Bid achieves competitive constraint satisfaction and value acquisition under sliding-window evaluation.
\end{abstract}

%%
%% The code below is generated by the tool at http://dl.acm.org/ccs.cfm.
%% Please copy and paste the code instead of the example below.
%%
\begin{CCSXML}
<ccs2012>
   <concept>
       <concept_id>10002951.10003227.10003447</concept_id>
       <concept_desc>Information systems~Computational advertising</concept_desc>
       <concept_significance>500</concept_significance>
       </concept>
 </ccs2012>
\end{CCSXML}

\ccsdesc[500]{Information systems~Computational advertising}

%%
%% Keywords. The author(s) should pick words that accurately describe
%% the work being presented. Separate the keywords with commas.
\keywords{Online Advertising, Auto-bidding, Sliding-Window Constraint}

% \received{20 February 2007}
% \received[revised]{12 March 2009}
% \received[accepted]{5 June 2009}

%%
%% This command processes the author and affiliation and title
%% information and builds the first part of the formatted document.
\maketitle

\footnotetext[1]{Work is done during the internship at Alibaba International Digital Commerce Group.}
\footnotetext[2]{Equal contribution.}

%% main text
% Introduction
\section{Introduction}

\begin{figure}[t]
    \centering
    \includegraphics[width=0.98\linewidth]{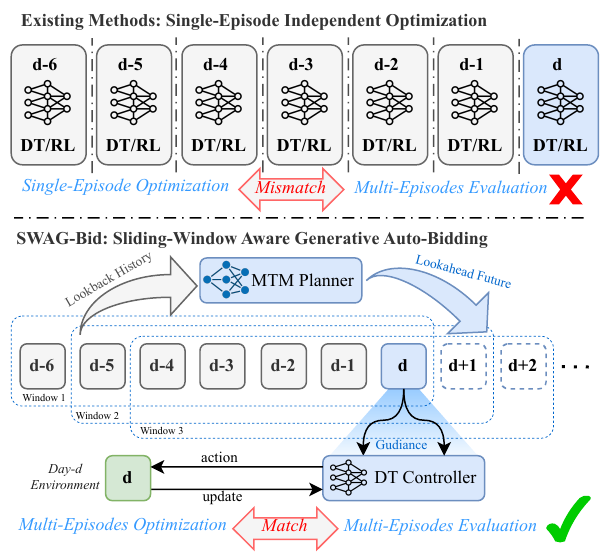}
    \vspace{-5pt}
    \caption{Single-episode optimization (\textbf{top}) vs. SWAG-Bid (\textbf{bottom}) under sliding-window constraints.}
    \vspace{-10pt}
    \label{fig:intro}
\end{figure}

Auto-bidding is the backbone of modern computational advertising~\cite{zhao2018deep, balseiro2021robust, ou2023deep}, handling billions of real-time bidding decisions every day. The task is formulated as constrained sequential decision-making, where an agent adjusts bids within a campaign period to maximize cumulative value~\cite{borissov2010automated} under ratio-based efficiency constraints such as Cost-Per-Action (CPA)~\cite{nazerzadeh2008dynamic, hu2016incentive}, Cost-Per-Click (CPC)~\cite{zhu2017optimized}, and Return on Ad Spend (ROAS)~\cite{deng2023multi}. To avoid the financial risks of online exploration, the community has turned to offline reinforcement learning~\cite{korenkevych2024offline} and generative sequence modeling, which learn from historical logs. Yet existing methods almost universally assume that \textit{each day is an independent optimization episode whose efficiency constraints are evaluated in isolation}. In practice, many advertisers produce value so sparsely that per-episode efficiency estimates become statistically meaningless. Such distorted metrics mislead advertisers into underestimating campaign effectiveness and reducing or halting spend, threatening long-term revenue stability and ecosystem health~\cite{yao2011dynamic, wilbur2009click}. Industrial platforms therefore evaluate efficiency constraints over sliding windows of typically $W{=}7$ episodes to sustain long-term advertising effectiveness~\cite{dekimpe1995persistence, sethuraman2011well}. This creates a fundamental gap between how methods optimize (per episode) and how performance is measured (across episodes), as illustrated in Figure~\ref{fig:intro}.

Sliding-window evaluation introduces \textbf{cross-episode coupling}, absent from the single-episode setting. The cost and value produced in episode $d$ enter the constraint computation of up to $W$ overlapping windows, so bidding decisions in episode $d$ must account for every future window containing it. Evaluating these effects requires anticipating market conditions over the next $W{-}1$ episodes. The problem therefore extends beyond a single-episode Markov Decision Process (MDP) into cross-episode sequential decision-making that requires forward-looking planning.

Existing auto-bidding methods, whether based on reinforcement learning~\cite{he2021unified, fujimoto2019off, kumar2020conservative, kostrikov2021offline} or generative sequence modeling~\cite{chen2021decision, guo2024generative, li2025gas, gao2025generative}, excel under the single-episode assumption but cannot handle cross-episode sliding-window constraints. A natural extension adds a Proportional-Integral-Derivative (PID) controller that adjusts per-episode targets based on historical window deviations. PID is purely reactive: it observes only the past $W{-}1$ episodes and cannot anticipate future market changes. This raises a core research question: \emph{How can an auto-bidding policy set per-episode bidding targets that balance value acquisition against all overlapping sliding-window constraints, when decisions in a single episode simultaneously affect up to $W$ windows and future market conditions are unknown?}

We propose \textbf{SWAG-Bid} (\textbf{S}liding-\textbf{W}indow \textbf{A}ware \textbf{G}enerative Auto-\textbf{Bid}ding), a hierarchical generative framework that splits cross-episode optimization into \textit{episode-level planning} and \textit{step-level execution}. The planner provides per-episode bidding guidance; the controller executes within each episode. Two challenges arise from this decomposition. \textbf{(1) How can the planner generate forward-looking targets that acquire value while respecting all affected sliding windows?} Inspired by Model Predictive Control (MPC), we adopt a Masked Trajectory Model (MTM) that serves as both policy and predictive model, using masked inference to predict market conditions in future episodes and generate candidate plans. We propose \textbf{Multi-Window MPC Sampling (MWMS)}, which scores these candidate plans against all affected sliding windows using exponential confidence decay. \textbf{(2) How can the planner convey its guidance to the controller in a step-adaptive manner?} Time steps within an episode face varying market conditions and remaining resources, so reliance on guidance should vary. We propose \textbf{Per-Step Gated Adaptive Layer Normalization (PSG-AdaLN)}, which conveys bidding-intensity signals through a per-step, state-adaptive gate trained under Classifier-Free Guidance (CFG), enabling the controller to adjust reliance on guidance at each step according to the current state. PSG-AdaLN, together with the Return-to-Go (RTG) and Cost-to-Go (CTG) channel that implicitly carries budget and efficiency constraint information, forms a dual-channel guidance mechanism. SWAG-Bid resolves cross-episode coupling through forward-looking planning rather than backward-looking reaction, with each component independently trainable.

Our contributions are as follows:
\begin{itemize}[leftmargin=*]
    \item We formalize auto-bidding under sliding-window efficiency constraints, identify the cross-episode coupling challenge, and propose SWAG-Bid, a hierarchical generative framework that decomposes the problem into episode-level proactive planning and step-level execution.
    \item We introduce MWMS for the episode-level planner: an MTM jointly acts as policy and predictive model through four-stage masked rollouts, predicts future market conditions to generate candidate plans, and scores them across all affected sliding windows with exponential confidence decay to select high-value targets with low violation risk.
    \item We introduce PSG-AdaLN for the step-level controller: a per-step, state-adaptive gate trained with CFG enables the controller to adjust reliance on planner guidance at each time step. Together with the RTG/CTG channel carrying budget and constraint information, this forms a dual-channel guidance mechanism.
    \item Experiments on AuctionNet-Sparse and online A/B tests on AliExpress show that SWAG-Bid achieves competitive performance in both constraint satisfaction and value acquisition.
\end{itemize}

\section{Preliminaries}

\subsection{Problem Formulation}

We follow the constrained auto-bidding formulation of~\cite{guo2024generative}. In the standard single-episode setting, an advertiser participates in $n$ impression auctions. Each impression $i$ carries a private valuation $v_i$, a potential cost $c_i$, and a value indicator $p_i$ that depends on the constraint type: conversion for CPA, click for CPC, or Gross Merchandise Volume (GMV) for ROAS. The objective is to maximize total value under a budget constraint $B$ and a ratio-based efficiency constraint $\rho_{\text{tgt}}$:
\begin{equation}
\label{eq:single_episode}
\begin{aligned}
\text{maximize} & \quad \textstyle\sum_{i=1}^{n} x_i v_i \\
\text{s.t.} & \quad \textstyle\sum_{i=1}^{n} c_ix_i \le B \\
& \quad \frac{\sum_{i=1}^{n} c_i x_i}{\sum_{i=1}^{n} p_i x_i} \le \rho_{\text{tgt}} \\
& \quad x_i \in \{0, 1\}, \quad \forall i
\end{aligned}
\end{equation}
where $x_i$ indicates whether impression $i$ is won and $\rho_{\text{tgt}}$ is the target efficiency ratio. By primal-dual theory~\cite{he2021unified}, when the optimization objective aligns with the constraint metric (e.g., maximizing conversions under CPA where $v_i = p_i$), the optimal bid reduces to a proportional strategy $b_i^* = \lambda^* v_i$, so the auto-bidding task becomes learning a policy $\pi$ that outputs the pacing parameter $\lambda_t$ at each step $t$. Existing methods~\cite{he2021unified, chen2021decision, guo2024generative} operate within this single-episode framework. In real-world systems, constraints are often evaluated over multiple episodes.

\textbf{Sliding-Window Constrained Auto-Bidding.}
We consider a setting closer to industry practice: \textit{the budget is independent per episode, while the efficiency constraint is evaluated over a $W$-episode sliding window.} Many advertisers have inherently sparse value; a single episode may yield minimal or no value, making per-episode efficiency estimates unreliable.

Consider an advertiser operating over $D$ consecutive episodes. In episode $d$, the advertiser participates in $n_d$ impression auctions, where each impression $i$ has private valuation $v_{d,i}$, potential cost $c_{d,i}$, value indicator $p_{d,i}$, and winning indicator $x_{d,i} \in \{0, 1\}$, with an independent budget $B_d$. As before, we assume the aligned setting $v_{d,i} = p_{d,i}$. The system uses proportional bidding $b_{d,i} = \lambda_{d,t} \cdot v_{d,i}$, with $T$ decision steps per episode. We define the aggregated episode-level cost and value as:
\begin{equation}
\mathcal{C}_d = \textstyle\sum_{i=1}^{n_d} c_{d,i} \, x_{d,i}, \quad \mathcal{R}_d = \textstyle\sum_{i=1}^{n_d} p_{d,i} \, x_{d,i}
\end{equation}

The complete optimization problem under sliding-window constraints is:

\begin{equation}
\label{eq:sliding_window_problem}
\begin{aligned}
\text{maximize} & \quad \textstyle\sum_{d=1}^{D} \sum_{i=1}^{n_d} x_{d,i} \, v_{d,i} \\
\text{s.t.} & \quad \textstyle\sum_{i=1}^{n_d} c_{d,i} \, x_{d,i} \leq B_d, \quad \forall d \in \{1, \ldots, D\} \\
& \quad \frac{\sum_{k=d-W+1}^{d} \mathcal{C}_k}{\sum_{k=d-W+1}^{d} \mathcal{R}_k} \leq \rho_{\text{tgt}}, \quad \forall d \geq W \\
& \quad x_{d,i} \in \{0, 1\}, \quad \forall d, i
\end{aligned}
\end{equation}

The first constraint enforces per-episode budget limits; the second requires the efficiency ratio over every $W$-episode sliding window to not exceed $\rho_{\text{tgt}}$. The fundamental difficulty lies in the \textbf{cross-episode coupling}: although budgets are decoupled across episodes, the efficiency constraint couples them, since decisions $\{x_{d,i}\}$ in episode $d$ simultaneously affect up to $W$ sliding windows. Thus, determining the optimal bidding strategy $\{\lambda_{d,t}\}_{t=1}^T$ for episode $d$ requires not only the intra-episode state but also anticipation of market conditions over the subsequent $W\!-\!1$ episodes. This extends the problem beyond a single-episode MDP to cross-episode sequential decision-making requiring proactive planning.

\subsection{Intra-Episode Bidding via DT}
\label{sec:dt_backbone_pre}

Following~\cite{wu2026constraint}, we formulate intra-episode auto-bidding as conditional sequence generation with constraint-decoupled dual goal streams. This Decision Transformer (DT) formulation separates value and cost conditioning, enabling the policy to perceive resource boundaries more effectively than standard RTG-only approaches. A trajectory is represented as:
\begin{equation}
\label{eq:trajectory}
\tau = \left( R_1, C_1, s_1, a_1, \dots, R_T, C_T, s_T, a_T \right)
\end{equation}
The trajectory components are:
\begin{itemize}[leftmargin=*]
    \item \textbf{State} $s_t$: intra-episode context encoding remaining time, remaining budget, and historical market features.
    \item \textbf{Action} $a_t = \lambda_t$: the bidding parameter at step $t$.
    \item \textbf{RTG} $R_t = \sum_{t'=t}^{T} r_{t'}$: target cumulative value from step $t$ onward.
    \item \textbf{CTG} $C_t = \sum_{t'=t}^{T} c_{t'}$: target cumulative cost from step $t$ onward.
\end{itemize}
The DT learns a goal-conditioned policy $\pi_\theta(a_t \mid R_{\leq t}, C_{\leq t}, s_{\leq t}, a_{<t})$. During inference, $R_1$ and $C_1$ are initialized to episode-level targets and updated recursively as outcomes are observed: $R_{t+1} = R_t - r_t$, $C_{t+1} = C_t - c_t$.

This goal-conditioning mechanism is well suited to \textbf{intra-episode pacing}: given a per-episode target, the DT allocates the budget across $T$ steps. Under the sliding-window setting of Eq.~\eqref{eq:sliding_window_problem}, a critical question arises: \textit{how should the episode-level targets $R_1$ and $C_1$ be determined?} The information boundary of the DT is confined to a single episode, as $s_t$ encodes only intra-episode state, with no visibility into the broader sliding window. The optimal values of $R_1$ and $C_1$ depend on the realized performance of neighboring episodes and anticipated market conditions in future episodes. The DT alone cannot perform this cross-episode target allocation, motivating a higher-level planning mechanism.

\section{Framework}
\label{sec:framework}

\begin{figure*}[th]
    \centering
    \includegraphics[width=0.99\linewidth]{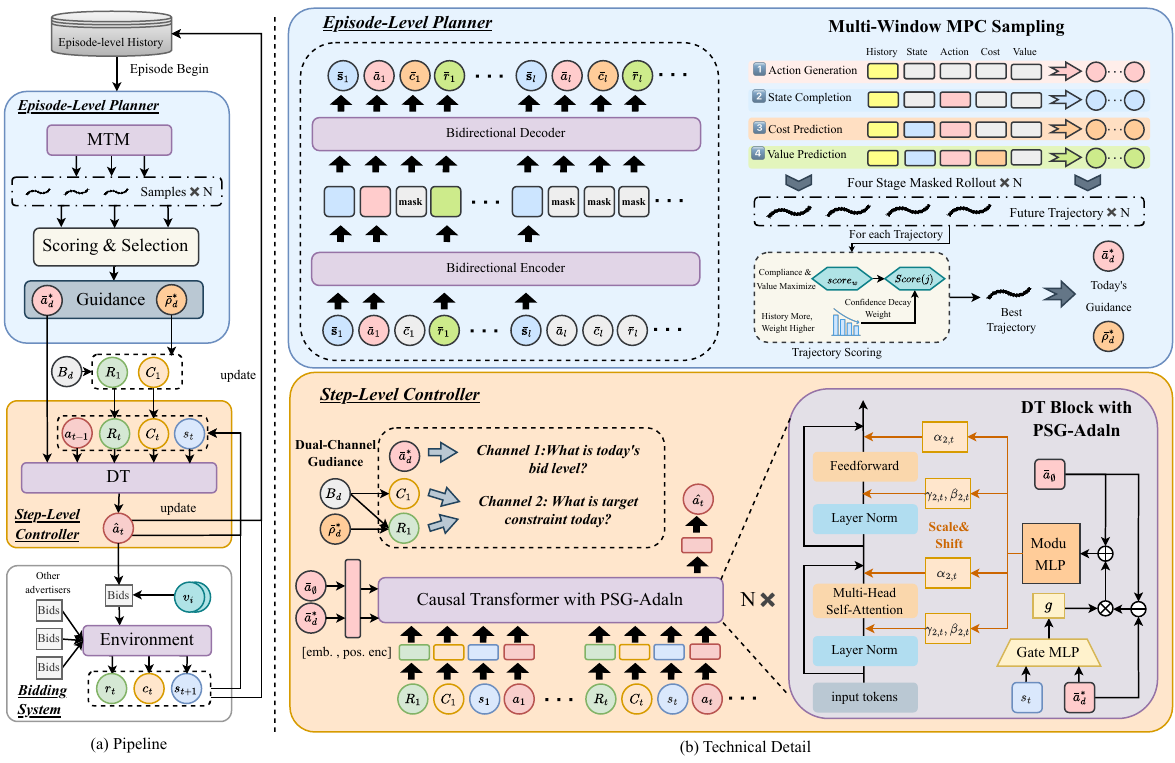}
    \vspace{-5pt}
    \caption{Overall framework of SWAG-Bid.}
    \vspace{-10pt}
    \label{fig:framework}
\end{figure*}

SWAG-Bid decomposes the cross-episode optimization of Eq.~\eqref{eq:sliding_window_problem} into two layers: an \textit{episode-level planner} (Sec.~\ref{sec:macro}) that provides bidding guidance via forward-looking planning over future sliding windows, and a \textit{step-level controller} (Sec.~\ref{sec:micro}) that executes bidding under this guidance. The planner uses an MTM as both policy and predictive model; the controller receives targets via the goal-conditioning interface of the DT and a state-adaptive guidance channel. At the start of each episode $d$, the planner generates and scores $N$ candidate trajectories via MWMS, extracting the action target $\bar{a}_d^*$ and market efficiency estimate $\bar{\rho}_d^*$. The controller receives guidance via a dual-channel mechanism: since the budget represents the advertiser's desired spend, the RTG/CTG channel anchors $C_1 = B_d$ and sets $R_1 = B_d/\bar{\rho}_d^*$ to carry predicted market efficiency, while the PSG-AdaLN channel injects $\bar{a}_d^*$. The controller executes $T$ steps, outputting $\lambda_{d,t}$ while recursively updating RTG/CTG tokens. Algorithm~\ref{alg:swag} (Appendix~\ref{app:algorithm}) formalizes the full procedure. Figure~\ref{fig:framework} shows the overall architecture.

\subsection{Episode-Level Planner}
\label{sec:macro}

The episode-level planner tackles cross-episode optimization: setting per-episode guidance pursuing high value under all affected sliding-window constraints. This requires generating candidate plans and predicting consequences. We unify both in one MTM.

\subsubsection{\textbf{Masked Trajectory Model}}
\label{sec:mtm}

The MTM operates on episode-level trajectory sequences $\tau^{\text{macro}} = \langle \bar{\mathbf{s}}_1, \bar{a}_1, \bar{c}_1, \bar{r}_1, \ldots, \bar{\mathbf{s}}_L, \bar{a}_L, \bar{c}_L, \bar{r}_L \rangle$, where $L$ covers both historical and future episodes. $\bar{\mathbf{s}}_d$ encodes market information, $\bar{a}_d$ is the episode mean bidding parameter, and $\bar{c}_d$, $\bar{r}_d$ denote episode-level cost and value. The architecture follows an MAE-style~\cite{he2022masked} Encoder-Decoder Transformer with modality-specific embeddings and sinusoidal positional encodings. The encoder processes only visible tokens, while the decoder reconstructs masked ones. Through different masking patterns, the MTM assumes dual roles: masking future actions lets it generate candidate plans (\textit{policy}), while masking future costs and values lets it predict their consequences (\textit{predictive model}).

\textbf{Training.}
A key challenge is that many historical episodes terminate early when the budget is exhausted, truncating observed cost and value $\mathcal{C}_d$ and $\mathcal{R}_d$. We rescale them to full-episode equivalents $\bar{c}_d$ and $\bar{r}_d$ using the historical intra-day traffic distribution, letting the MTM learn the potential market output rather than budget-limited observations. During training, we adopt a randomized masking strategy~\cite{wu2023masked}, where each sample selects mask ratios and truncation points, exposing the model to diverse masking patterns that unify policy and predictive model roles. We further employ constraint-aware data curation: training windows with frequent constraint violations are filtered out, and retained samples are weighted by a constraint-aware quality score. We define a penalty function $\phi(\rho, \rho_{\text{tgt}}) = \min\!\left((\rho_{\text{tgt}}/\rho)^q,\; 1\right)$, and assign each sample $\tau$ a weight:
\begin{equation}
\label{eq:awr_weight}
u_\tau = \frac{1}{|\mathcal{W}_\tau|} \sum_{w \in \mathcal{W}_\tau} \phi\!\left(\frac{\sum_{d' \in w} \mathcal{C}_{d'}}{\sum_{d' \in w} \mathcal{R}_{d'}},\; \rho_{\text{tgt}}\right) \cdot \sum_{d' \in w} \mathcal{R}_{d'}
\end{equation}
where $\mathcal{W}_\tau$ denotes the sliding windows overlapping $\tau$.
The MTM is trained with a weighted masked reconstruction loss. Deterministic modalities (states, costs, values) use Mean Squared Error (MSE), while the action modality is modeled with a stochastic head:
\begin{equation}
\label{eq:macro_loss}
\mathcal{L}_{\text{macro}} = -\mathbb{E}_{\tau}\! \left[\sum_{(d,m) \in \mathcal{M}} u_\tau \cdot \log P_\theta\!\left(z_d^m \mid \text{Masked}(\tau)\right) \right]
\end{equation}
where $z_d^m$ denotes the token of modality $m$ at episode $d$, and $\mathcal{M}$ is the set of masked positions. The action head incorporates entropy regularization~\cite{haarnoja2018soft} to maintain sampling diversity.

\subsubsection{\textbf{MWMS: Multi-Window MPC Sampling}}
\label{sec:mwms}

At inference time, the planner must evaluate how decisions in the current episode affect all $W$ sliding windows containing it. MWMS uses two steps: generating candidate trajectories via four-stage masked rollouts, and selecting the best via multi-window constraint-aware scoring.

\textbf{Four-Stage Masked Rollout.}
Given the current episode $d$ and the completed historical episodes, the MTM generates $N$ complete trajectory rollouts through four sequential stages:
\begin{itemize}[leftmargin=*]
    \item \textbf{Stage 1 (Action Generation).} Masking future actions, the MTM acts as a policy and samples $N$ candidate action sequences for the planning horizon.
    \item \textbf{Stage 2 (State Completion).} Conditioned on the sampled actions, the MTM predicts future episode states.
    \item \textbf{Stage 3 (Cost Prediction).} Conditioned on the known actions and states, the MTM predicts future episode costs.
    \item \textbf{Stage 4 (Value Prediction).} Conditioned on the known actions, states, and costs, the MTM predicts future episode values, completing the trajectory.
\end{itemize}
Each stage conditions on all information revealed by previous stages and requires a single batched forward pass ($N$ samples in parallel), yielding $N$ complete trajectories in 4 forward passes.

\textbf{Multi-Window Constraint Scoring.}
Since predictions inherit the rescaled training space, each episode's predicted cost is clamped at the budget with value scaled proportionally, restoring attainable levels. Each trajectory is scored by jointly evaluating all $W$ sliding windows including the current episode. For window $w$, its efficiency ratio is $\rho_w = \frac{\sum_{d' \in w} \mathcal{C}_{d'}}{\sum_{d' \in w} \mathcal{R}_{d'}}$, and the window score uses the penalty $\phi$ from training, softly enforcing the window constraint of Eq.~\eqref{eq:sliding_window_problem}:
\begin{equation}
\label{eq:window_score}
\text{score}_{w} = \phi(\rho_w, \rho_{\text{tgt}}) \cdot \sum_{d' \in w} \mathcal{R}_{d'}
\end{equation}
The total score for trajectory $j$ is a weighted sum across all windows:
\begin{equation}
\label{eq:total_score}
\text{Score}(j) = \sum_{w} \zeta_w \cdot \text{score}_{w}^{(j)}
\end{equation}
where the weight $\zeta_w$ uses exponential confidence decay to reflect the decreasing reliability of model predictions over longer horizons:
\begin{equation}
\label{eq:exp_decay}
\zeta_w = \exp(-\kappa \cdot f_w), \quad f_w = \frac{|\{d' \in w : d' \geq d\}|}{|w|}
\end{equation}
Here $f_w$ is the fraction of predicted episodes in window $w$. Near-term windows with mostly observed data receive higher weights, while far-term windows dominated by predictions are down-weighted. The trajectory with the highest score is selected, from which we extract the action target $\bar{a}_d^*$ and market efficiency estimate $\bar{\rho}_d^* = \mathcal{C}_d^*/\mathcal{R}_d^*$ for the current episode as controller guidance.

\subsection{Step-Level Controller}
\label{sec:micro}

The step-level controller executes intra-episode bidding under guidance from the planner. This guidance is transmitted through a dual-channel mechanism: the RTG/CTG channel implicitly carries budget and efficiency constraint information, while the PSG-AdaLN channel carries bidding-intensity signals. Since this guidance is an episode-level signal, yet intra-episode steps face varying market conditions and remaining resources, the controller must adaptively adjust its reliance on this guidance at each step.

\subsubsection{\textbf{Guidance-Conditioned Decision Transformer}}
\label{sec:dt_backbone}

The step-level controller instantiates the constraint-decoupled DT described in Sec.~\ref{sec:dt_backbone_pre} as a causal Transformer with a Diagonal Gaussian action head~\cite{zheng2022online}. The training objective is:
\begin{equation}
\label{eq:micro_loss}
\mathcal{L}_{\text{micro}} = -\mathbb{E}\left[\log \pi_\theta(a_t \mid R_{\leq t}, C_{\leq t}, s_{\leq t}, a_{<t}) + \eta \cdot \mathcal{H}(\pi_\theta)\right]
\end{equation}
where $\mathcal{H}(\pi_\theta)$ is the policy entropy and $\eta$ is an automatically tuned temperature parameter. During training, RTG/CTG tokens use ground-truth values from data, and the action target is set to the ground-truth episode mean bid (\textit{oracle}) as input to PSG-AdaLN.

\subsubsection{\textbf{PSG-AdaLN: Per-Step Gated Adaptive Layer Normalization}}
\label{sec:psg}

The action target is an episode-level signal, yet intra-episode steps face varying market states and remaining budgets, calling for varying reliance on the guidance. Planner predictions also inevitably carry estimation errors, necessitating a mechanism to adaptively modulate reliance at each step. This per-step compliance should be learned from data rather than imposed as a fixed temporal pattern. PSG-AdaLN addresses these requirements by extending DiT-style AdaLN~\cite{peebles2023scalable} with per-step state-adaptive gating.

\textbf{Guidance Encoding.}
The action target $\bar{a}_d^*$ is mapped to a global conditioning vector shared across all steps within the episode:
\begin{equation}
\label{eq:guidance_enc}
\mathbf{g} = \text{MLP}_{\text{guide}}(\bar{a}_d^*) \in \mathbb{R}^{H}
\end{equation}

\textbf{Classifier-Free Guidance Training.}
During training, the action target is dropped with probability $p_{\text{drop}}$, enabling the model to learn distinct behavioral modes for guided ($\mathbf{g}$) and unguided ($\mathbf{g}_\emptyset$) conditions. This dual-mode design is a prerequisite for the per-step gating mechanism, allowing interpolation between the two modes. Retained targets are further perturbed with bounded multiplicative noise, exposing the gate to imperfect guidance. At inference time, the oracle target is replaced by the planner prediction.

\textbf{Per-Step Gating.}
Building on these two modes, PSG-AdaLN introduces a per-step gate that determines guidance compliance from intra-episode context. At each step $t$, the controller extracts the state token $\mathbf{x}_{s,t}$ and computes a step-specific gate:
\begin{equation}
\label{eq:gate}
g_t = \sigma\!\left(\text{MLP}_{\text{gate}}\!\left([\mathbf{g};\; \mathbf{x}_{s,t}]\right)\right) \in [0, 1]
\end{equation}
The gate interpolates between conditional and unconditional:
\begin{equation}
\label{eq:gate_interp}
\tilde{\mathbf{g}}_t = \mathbf{g}_\emptyset + g_t \cdot (\mathbf{g} - \mathbf{g}_\emptyset)
\end{equation}

\textbf{AdaLN Modulation.}
The gated embedding $\tilde{\mathbf{g}}_t$ is injected into each Transformer block via AdaLN modulation, producing scale $\gamma$, shift $\beta$, and gate $\alpha$ modulation vectors for the attention and MLP sub-layers:
\begin{align}
\label{eq:adaln_attn}
\mathbf{h}_t &= (1 + \gamma_{1,t}) \odot \text{LN}(\mathbf{x}_t) + \beta_{1,t} \\
\mathbf{x}_t &\leftarrow \mathbf{x}_t + \alpha_{1,t} \odot \text{Attn}(\mathbf{h}_t) \\
\label{eq:adaln_mlp}
\mathbf{h}_t &= (1 + \gamma_{2,t}) \odot \text{LN}(\mathbf{x}_t) + \beta_{2,t} \\
\mathbf{x}_t &\leftarrow \mathbf{x}_t + \alpha_{2,t} \odot \text{MLP}(\mathbf{h}_t)
\end{align}
The gate weights $\alpha$ are initialized to zero, so that at the start of training the guidance has no effect and the model is equivalent to a standard DT. The gates progressively open only when the guidance signal tends to improve action prediction.

% Experiments

\newcolumntype{C}{>{\Centering\arraybackslash}X}

\begin{table*}[ht]
\centering
\caption{Performance comparison under SW-Score ($\uparrow$, higher is better) and SW-ER ($\downarrow$, lower is better) across different budget settings on AuctionNet-Sparse. The boldface denotes the best result; the underline indicates the second-best result. ``*'' indicates statistically significant improvements over the best baseline (paired two-sided $t$-test across advertisers, $p<0.05$).}
\label{tab:main_table}
\vspace{-10pt}
\renewcommand{\arraystretch}{1.2}
\resizebox{0.98\textwidth}{!}{
\begin{tabularx}{\textwidth}{ c | c | *{7}{C} c c c }
\hline\hline
Budget & Metric & TD3-BC & BCQ & CQL & IQL & DT & CDT & GAS & PRO-Bid & PID+DT & SWAG-Bid \\
\hline
\multirow{2}{*}{50\%}  & SW-Score $\uparrow$ & 83.7  & 109.5 & 99.6  & 117.2 & 93.6  & 119.9 & 115.2 & $\underline{124.3}$ & 113.8 & \textbf{129.7*} \\
                       & SW-ER $\downarrow$  & 0.265 & 0.336 & 0.345 & \underline{0.256} & 0.470 & 0.307 & 0.381 & 0.274 & 0.313 & $\textbf{0.244*}$     \\
\cline{1-12}
\multirow{2}{*}{75\%}  & SW-Score $\uparrow$ & 113.0 & 144.0 & 136.4 & 152.4 & 143.0 & 154.5 & 162.4 & $\underline{174.0}$ & 163.0 & \textbf{185.5*} \\
                       & SW-ER $\downarrow$  & 0.271 & 0.321 & 0.393 & $\underline{0.250}$ & 0.435 & 0.330 & 0.366 & 0.298 & 0.327 & \textbf{0.235*}     \\
\cline{1-12}
\multirow{2}{*}{100\%} & SW-Score $\uparrow$ & 147.8 & 173.3 & 167.5 & 180.5 & 191.0 & 193.2 & 207.4 & $\underline{227.1}$ & 205.8 & \textbf{240.7*} \\
                       & SW-ER $\downarrow$  & 0.277 & 0.310 & 0.399 & $\underline{0.232}$ & 0.443 & 0.315 & 0.366 & 0.283 & 0.336 & \textbf{0.208*}     \\
\cline{1-12}
\multirow{2}{*}{125\%} & SW-Score $\uparrow$ & 184.5 & 204.2 & 190.8 & 201.3 & 237.7 & 220.1 & 249.3 & $\underline{262.9}$ & 241.1 & \textbf{276.5*} \\
                       & SW-ER $\downarrow$  & 0.292 & 0.283 & 0.438 & $\underline{0.259}$ & 0.402 & 0.296 & 0.339 & 0.289 & 0.360 & \textbf{0.229*}     \\
\cline{1-12}
\multirow{2}{*}{150\%} & SW-Score $\uparrow$ & 213.3 & 222.7 & 213.3 & 217.0 & 281.4 & 239.4 & 295.5 & $\underline{303.1}$ & 284.3 & \textbf{315.4*} \\
                       & SW-ER $\downarrow$  & 0.313 & 0.289 & 0.455 & 0.280 & 0.435 & $\underline{0.271}$ & 0.351 & 0.277 & 0.354 & \textbf{0.253*}     \\
\hline\hline
\end{tabularx}}
\vspace{-5pt}
\end{table*}

\section{Experiments}

In this section, we conduct offline and online experiments to address the following research questions:

\begin{itemize}[leftmargin=*]
    \item \textbf{RQ1:} \emph{Does SWAG-Bid outperform state-of-the-art baselines in both value maximization and sliding-window constraint satisfaction?}
    \item \textbf{RQ2:} \emph{How does proactive planning compare against reactive control for cross-episode constraint management?}
    \item \textbf{RQ3:} \emph{How do the key components individually contribute to overall performance?}
    \item \textbf{RQ4:} \emph{Does the MWMS sampling-and-selection mechanism improve plan quality over deterministic prediction?}
    \item \textbf{RQ5:} \emph{Does the PSG-AdaLN gate exhibit awareness of both intra-episode time step and guidance quality?}
    \item \textbf{RQ6:} \emph{Can SWAG-Bid be effectively adapted to real-world online advertising environments?}
\end{itemize}

\subsection{Experimental Setup}

\subsubsection{\textbf{Dataset}}

We conduct offline experiments on AuctionNet-Sparse~\cite{su2024auctionnet}, the sparse-conversion variant of the AuctionNet benchmark. We choose this variant rather than the dense subset because the core motivation of our work is that many advertisers have inherently sparse value, making per-episode efficiency estimates unreliable. AuctionNet-Sparse, with its low conversion rates and wide CPA ranges, naturally reflects this real-world challenge. Models are trained on the historical bidding logs of AuctionNet-Sparse.

\textbf{Multi-Episode Continuous Evaluation.}
The standard AuctionNet benchmark provides only a single-episode evaluation protocol, which is insufficient for assessing sliding-window constraint satisfaction. To enable multi-episode evaluation, we use the sparse multi-agent bidding simulator provided by the AuctionNet benchmark to generate $D = 21$ consecutive episodes of bidding logs. Each episode corresponds to one day with an independent budget and $T = 48$ decision steps. The efficiency constraint is evaluated over $W = 7$-episode sliding windows. We evaluate 48 advertisers with diverse budget and CPA configurations in parallel. We report metrics averaged across all advertiser--episode pairs.

\subsubsection{\textbf{Evaluation Metrics}}

Standard single-episode metrics are insufficient for the sliding-window setting, as they do not capture cross-episode constraint dynamics. We extend the evaluation framework with window-level metrics:

\begin{itemize}[leftmargin=*]
    \item \textbf{SW-Score (Sliding-Window Score)}: The primary metric. For a sliding window $w$, let $\rho_w = {\sum_{d \in w} \mathcal{C}_d}/{\sum_{d \in w} \mathcal{R}_d}$ be its efficiency ratio. The SW-Score is:
    \begin{equation}
    \label{eq:sw_score}
    \text{SW-Score} = \min\!\left(\left(\frac{\rho_{\text{tgt}}}{\rho_w}\right)^{\!q},\; 1\right) \cdot \sum_{d \in w} \mathcal{R}_d
    \end{equation}
    where $q = 2$. The penalty is active when $\rho_w > \rho_{\text{tgt}}$ and discounts the window reward proportional to the violation severity; otherwise the full reward is retained. The reported value is averaged across all advertiser--episode pairs.
    \item \textbf{SW-ER (Sliding-Window Exceed Rate)}: The fraction of sliding windows where the efficiency constraint is violated ($\rho_w > \rho_{\text{tgt}}$), which is independent of the penalty form. Lower SW-ER indicates more reliable constraint satisfaction.
\end{itemize}

\subsubsection{\textbf{Baselines}}

Since all existing auto-bidding methods are designed for the single-episode setting, we extend them to the multi-episode scenario. Offline RL methods include TD3-BC~\cite{fujimoto2021minimalist}, which adds behavioral cloning regularization to constrain the policy to the behavior data support; BCQ~\cite{fujimoto2019off}, which constrains Q-learning to remain close to the behavior policy; CQL~\cite{kumar2020conservative}, which learns a conservative Q-function to penalize out-of-distribution actions; and IQL~\cite{kostrikov2021offline}, which avoids explicit behavior cloning via implicit in-sample value functions. Generative methods include DT~\cite{chen2021decision}, which employs a transformer architecture for sequential decision-making with fixed RTG/CTG initialization and no cross-episode information; CDT~\cite{liu2023constrained}, a DT-based approach that handles multiple constraints via explicit constraint tokens; GAS~\cite{li2025gas}, a DT-based framework with post-training optimization via Monte Carlo Tree Search; and PRO-Bid~\cite{wu2026constraint}, which balances value and constraint adherence through Pareto-prioritized regret optimization. PID+DT serves as the only cross-episode baseline, where a PID controller adjusts per-episode RTG/CTG targets based on the window efficiency deviation of the historical $W\!-\!1$ episodes, with DT handling intra-episode pacing, but its adjustment is purely reactive. All single-episode methods treat each of the 21 episodes independently with fixed initialization. Implementation details are in Appendix~\ref{app:impl}.

\subsection{Overall Performance (RQ1)}
\label{sec:rq1}

Table~\ref{tab:main_table} reports the SW-Score and SW-ER of SWAG-Bid and nine competitive baselines across five budget settings on AuctionNet-Sparse.

\textbf{SWAG-Bid achieves the best empirical trade-off between value maximization and constraint satisfaction among compared methods.} Offline RL methods scale poorly with budget, as conservative value estimation limits their use of additional budget. Generative methods exploit budget more effectively but lack cross-episode visibility, leading to frequent sliding-window violations. SWAG-Bid bridges this gap through forward-looking planning, achieving the highest SW-Score and lowest SW-ER across all budgets. PRO-Bid ranks second in SW-Score, suggesting that explicit constraint-aware optimization aids value acquisition but cannot fully compensate for the absence of cross-episode planning.

\textbf{Cross-episode information is beneficial for sliding-window constraint satisfaction.} PID+DT, the only cross-episode baseline, achieves lower SW-ER than its single-episode counterpart DT across all budgets, and also leads in SW-Score, with the SW-Score gap most pronounced at low budgets but narrowing as budget increases. This suggests reactive cross-episode correction helps primarily when budget is tight, but its benefit diminishes as budget loosens. PID+DT still lags significantly behind SWAG-Bid, confirming that effective constraint satisfaction requires proactively anticipating future market dynamics rather than reactively correcting past deviations.

\subsection{Proactive vs Reactive Planning (RQ2)}
\label{sec:rq2}

To isolate whether the improvement of SWAG-Bid stems from forward-looking prediction or merely more sophisticated scoring, we design two variants: PID+Controller replaces the SWAG-Bid planner with a PID controller, retaining the step-level controller, and SWAG-Bid-w/o-Future retains MWMS scoring but removes future episode prediction, making it reactive while preserving the scoring framework. Table~\ref{tab:rq2_proactive_vs_reactive} summarizes results at 100\% budget.

\textbf{Forward-looking prediction primarily drives constraint satisfaction.} While both scoring and prediction contribute to performance, their effects differ across metrics: MWMS scoring improves SW-Score but has limited impact on SW-ER, whereas adding future prediction yields a much larger improvement in SW-ER, roughly double that of scoring alone. This indicates that multi-window scoring helps identify higher-value trajectories, but the ability to predict future episodes is the primary driver of proactive constraint management, as it enables the planner to anticipate and avoid potential violations before they occur.

\begin{table}[htbp]
\centering
\caption{Proactive vs.\ reactive planning at 100\% budget.}
\label{tab:rq2_proactive_vs_reactive}
\vspace{-10pt}
\renewcommand{\arraystretch}{1.15}
\small
\begin{tabular}{lcc}
\toprule
Method & SW-Score $\uparrow$ & SW-ER $\downarrow$ \\
\midrule
PID+Controller       & 220.9 & 0.289 \\
SWAG-Bid-w/o-Future     & 231.3 & 0.262 \\
\textbf{SWAG-Bid (Full)}& \textbf{240.7} & \textbf{0.208} \\
\bottomrule
\end{tabular}
\vspace{-10pt}
\end{table}

\subsection{Ablation Study (RQ3)}
\label{sec:rq3}

To quantify the contribution of each component, we systematically remove the Planner, MWMS, and PSG-AdaLN from the full SWAG-Bid framework. Table~\ref{tab:rq3_ablation} reports the ablation results at 100\% budget.

\textbf{The Planner is the most impactful component.} Removing it causes the most severe degradation in both metrics, suggesting that hierarchical planning provides the foundation for cross-episode constraint satisfaction; without it, the controller operates with fixed targets and loses all visibility into future market dynamics.

\textbf{MWMS scoring primarily improves value acquisition.} Replacing MWMS with single-window scoring causes a notable drop in SW-Score but the smallest increase in SW-ER among all ablations. This asymmetry shows that multi-window evaluation helps identify higher-value trajectories without meaningfully compromising constraint compliance, as the per-window penalty in the scoring function already ensures basic constraint awareness.

\textbf{PSG-AdaLN is important for constraint satisfaction.} Removing it causes the second-largest degradation in both metrics, with SW-ER rising sharply, suggesting that per-step adaptive guidance helps the controller execute plans more reliably. A comparison against alternative guidance injection methods is provided in Table~\ref{tab:conditioning} (Appendix~\ref{app:conditioning}).

\begin{table}[htbp]
\centering
\caption{Ablation study at 100\% budget.}
\label{tab:rq3_ablation}
\vspace{-10pt}
\renewcommand{\arraystretch}{1.15}
\small
\begin{tabular}{lcc}
\toprule
Variant & SW-Score $\uparrow$ & SW-ER $\downarrow$ \\
\midrule
w/o Planner            & 216.3 & 0.336 \\
w/o MWMS               & 231.0 & 0.256 \\
w/o PSG-AdaLN          & 227.1 & 0.327 \\
\textbf{SWAG-Bid (Full)}    & \textbf{240.7} & \textbf{0.208} \\
\bottomrule
\end{tabular}
\vspace{-10pt}
\end{table}

\subsection{MWMS Mechanism Analysis (RQ4)}
\label{sec:rq4}

\begin{figure}[htbp]
\centering
\includegraphics[width=0.75\linewidth]{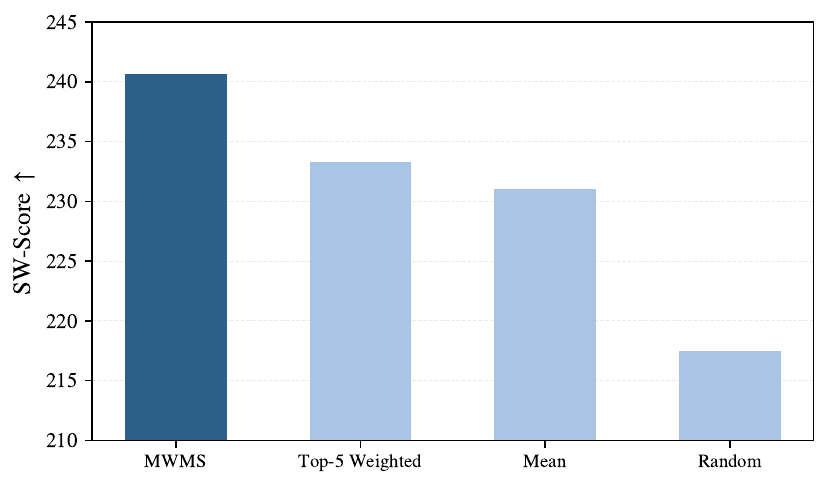}
\vspace{-10pt}
\caption{Four selection strategies compared at 100\% budget.}
\label{fig:rq4_mwms_selection}
\vspace{-10pt}
\end{figure}

To verify whether the MWMS sampling-and-selection mechanism improves plan quality over deterministic prediction, we compare four selection strategies. Mean deterministically outputs a single trajectory. Random uniformly picks one candidate from the sampled pool without scoring. Top-5 Weighted computes a softmax-weighted average of the top-5 candidates. MWMS selects the candidate with the highest multi-window score. Figure~\ref{fig:rq4_mwms_selection} compares the four strategies at 100\% budget.

\textbf{MWMS scoring helps identify higher-quality trajectories.} Random falling below Mean shows that sampling without a quality signal is harmful, as unguided selection introduces noise rather than value. Top-5 Weighted outperforming Mean indicates that aggregating multiple candidates smooths individual prediction errors, yet averaging inevitably blurs the sharpest trajectory. MWMS outperforms both by directly selecting the highest-scoring candidate rather than aggregating, preserving the full constraint-aware signal that aggregation would dilute. This suggests that in the presence of prediction uncertainty, selecting a single high-quality plan is more effective than averaging across plans of varying quality.

\subsection{PSG-AdaLN Gate Awareness (RQ5)}
\label{sec:rq5}

\begin{figure*}[htbp]
\centering
\includegraphics[width=0.98\linewidth]{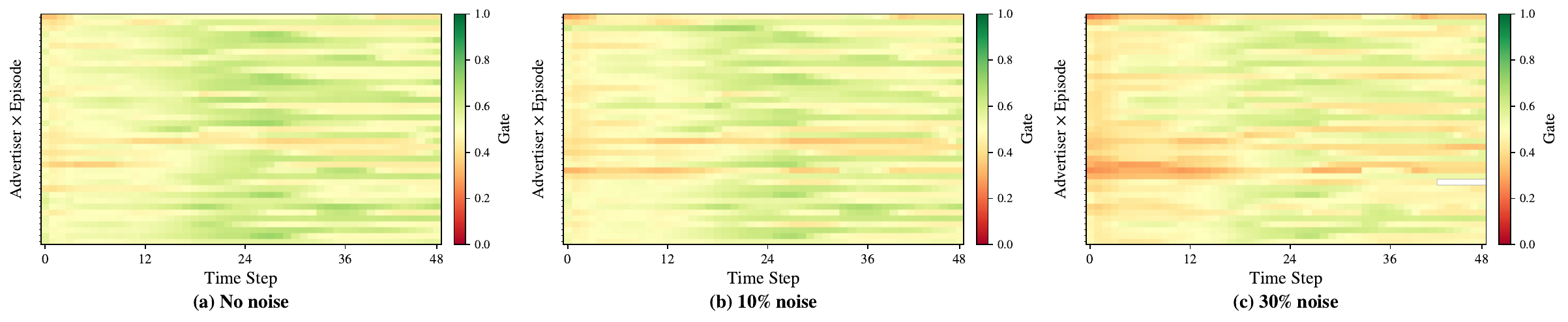}
\vspace{-10pt}
\caption{Per-step gate values under varying guidance noise. Each panel shows the Time Step heatmap for a randomly sampled subset of advertiser--episode pairs.}
\label{fig:gate_heatmap}
\vspace{-5pt}
\end{figure*}

To verify that the PSG-AdaLN gate is aware of both intra-episode time step and guidance quality, we inject varying levels of multiplicative noise into the action target at inference time and record the layer-mean gate value at every time step (Figure~\ref{fig:gate_heatmap}).

\textbf{Time-step awareness.} Within each panel, gate values vary across time steps rather than remaining constant, suggesting that the per-step gate dynamically adjusts guidance compliance based on intra-episode state. The variation is also visible across rows, indicating the gate adapts to different advertiser--episode contexts rather than degenerating into a fixed constant.

\textbf{Guidance-quality awareness.} Across panels, gate values decrease as noise increases. Without noise, the gate stays high, indicating trust in planner guidance. As noise grows, the gate reduces its magnitude, suggesting adaptive down-weighting under corrupted guidance. This quality-aware gating provides a degree of robustness to degraded guidance, as quantified in Appendix~\ref{app:noise_robustness}.

\subsection{Online A/B Test (RQ6)}
\label{sec:rq6}

To evaluate the practical effectiveness of SWAG-Bid, we deployed it on the advertising system of AliExpress, a large-scale cross-border e-commerce platform, where advertisers specify a per-episode budget and a ROAS efficiency constraint evaluated over 7-day sliding windows. We compared SWAG-Bid against the production baseline, a single-episode DT-based policy, across a wide range of advertising categories. Adaptation and deployment details are provided in Appendix~\ref{app:deploy}.

Online A/B testing was conducted over a consecutive 21-day period. Randomization was performed at the campaign level: eligible campaigns were randomly assigned to the two groups, with 10\% of traffic allocated to the experimental SWAG-Bid group; exact campaign counts are withheld due to company policy. The results are summarized in Table~\ref{tab:online_results}, where each metric is a campaign-level average and its lift is computed via difference-in-differences between the SWAG-Bid and control groups, and Achievement Rate denotes the fraction of campaigns satisfying the sliding-window constraint. SWAG-Bid increases average campaign cost by 1.96\% and GMV by 3.42\%, improving ROAS by 5.65\%: advertisers spend more on the platform while obtaining better returns. These gains are accompanied by a 2.02pp improvement in the constraint achievement rate. This suggests that SWAG-Bid can be adapted from the offline CPA setting to the online ROAS constraint type, and its forward-looking planning mechanism translates to improvements in cross-episode constraint satisfaction under real-world market dynamics.

\begin{table}[htbp]
\centering
\caption{Online A/B test results.}
\label{tab:online_results}
\vspace{-5pt}
\small
\begin{tabular}{lcccc}
\toprule
Metric & Cost & GMV & ROAS & Achievement Rate \\
\midrule
Improve & +1.96\% & +3.42\% & +5.65\% & +2.02pp \\
\bottomrule
\end{tabular}
\vspace{-5pt}
\end{table}

% Related Work
\section{Related Work}

\subsection{Auto-Bidding}
Auto-bidding optimizes key performance indicators under strict budget and efficiency constraints~\cite{deng2021towards,aggarwal2024auto}. Early control-theoretic approaches such as PID~\cite{chen2011real} and OnlineLP~\cite{yu2017online} achieve efficient pacing but rely on simplified assumptions. Deep RL methods, including RLB~\cite{cai2017real}, USCB~\cite{he2021unified}, MAAB~\cite{wen2022cooperative}, and SORL~\cite{mou2022sustainable}, handle high-dimensional states, with the industry later shifting to offline RL frameworks such as BCQ~\cite{fujimoto2019off}, CQL~\cite{kumar2020conservative}, and IQL~\cite{kostrikov2021offline} to mitigate online exploration risks. The MDP assumption inherent in these methods limits their ability to capture long-horizon dependencies, motivating generative auto-bidding: DiffBid~\cite{guo2024generative} and EGDB~\cite{peng2025expert} generate complete trajectories via conditional denoising, while DT extensions enrich the conditioning interface; CDT~\cite{liu2023constrained} introduces constraint tokens, GAS~\cite{li2025gas} applies post-training MCTS, GAVE~\cite{gao2025generative} adopts value-guided exploration, EBaReT~\cite{li2025ebaret} incorporates expert-guided inference, and PRO-Bid~\cite{wu2026constraint} performs Pareto-prioritized regret optimization. These methods all operate within a single-episode framework with no cross-episode visibility. Our work fills this gap by introducing episode-level forward-looking planning that explicitly anticipates cross-episode constraint coupling.

\subsection{Generative Sequence Modeling}
Generative sequence modeling reformulates sequential decision making as conditional token prediction, enabling a single Transformer to capture long-range dependencies that elude bootstrapped value learning. This paradigm originates from two complementary perspectives: Trajectory Transformer~\cite{janner2021offline} models offline RL as autoregressive sequence generation with beam-search planning, while Decision Transformer~\cite{chen2021decision} simplifies it to goal-conditioned action generation via return-to-go prompting. The autoregressive formulation was generalized by masked prediction, including the Masked Trajectory Model~\cite{wu2023masked}, UniMASK~\cite{carroll2022uni}, MaskDP~\cite{liu2022masked}, and CurrMask~\cite{tang2024learning}, which train on randomly masked token subsets so one network can serve as policy, dynamics model, or value estimator depending on the inference-time masking pattern. As an alternative generative interface, diffusion-based planners such as Diffuser~\cite{janner2022planning} and Decision Diffuser~\cite{ajay2022conditional} formulate trajectory optimization as iterative denoising. Architecturally, DiT~\cite{peebles2023scalable} introduces AdaLN modulation, providing conditioning control for Transformer-based generation. Our work applies the masked-trajectory paradigm at the episode level for cross-episode constraint planning, extending its use beyond single-agent robotics control to window-aggregated efficiency constraints.

% Conclusion
\section{Conclusion}

In this paper, we propose SWAG-Bid, a hierarchical generative auto-bidding framework for cross-episode constraint coupling under sliding-window evaluation. SWAG-Bid decomposes the problem into episode-level planning via MWMS, which samples and scores candidate plans across all affected sliding windows, and step-level execution via PSG-AdaLN, which adaptively modulates guidance compliance at each step. Offline experiments on AuctionNet-Sparse show that SWAG-Bid achieves the best empirical trade-off between value maximization and constraint satisfaction among compared methods. Online A/B testing on AliExpress further suggests its adaptability to real-world ROAS-constraint settings. By enabling forward-looking cross-episode planning, SWAG-Bid closes the gap between per-episode optimization and window-level evaluation.

%%
%% The next two lines define the bibliography style to be used, and
%% the bibliography file.
\bibliographystyle{ACM-Reference-Format}
\bibliography{chapter/8ref}

@article{su2024auctionnet,
  title={Auctionnet: A novel benchmark for decision-making in large-scale games},
  author={Su, Kefan and Huo, Yusen and Zhang, Zhilin and Dou, Shuai and Yu, Chuan and Xu, Jian and Lu, Zongqing and Zheng, Bo},
  journal={Advances in Neural Information Processing Systems},
  volume={37},
  pages={94428--94452},
  year={2024}
}

@inproceedings{fujimoto2019off,
  title={Off-policy deep reinforcement learning without exploration},
  author={Fujimoto, Scott and Meger, David and Precup, Doina},
  booktitle={International conference on machine learning},
  pages={2052--2062},
  year={2019},
  organization={PMLR}
}

@article{kumar2020conservative,
  title={Conservative q-learning for offline reinforcement learning},
  author={Kumar, Aviral and Zhou, Aurick and Tucker, George and Levine, Sergey},
  journal={Advances in neural information processing systems},
  volume={33},
  pages={1179--1191},
  year={2020}
}

@article{kostrikov2021offline,
  title={Offline reinforcement learning with implicit q-learning},
  author={Kostrikov, Ilya and Nair, Ashvin and Levine, Sergey},
  journal={arXiv preprint arXiv:2110.06169},
  year={2021}
}

@inproceedings{guo2024generative,
  title={Generative auto-bidding via conditional diffusion modeling},
  author={Guo, Jiayan and Huo, Yusen and Zhang, Zhilin and Wang, Tianyu and Yu, Chuan and Xu, Jian and Zheng, Bo and Zhang, Yan},
  booktitle={Proceedings of the 30th ACM SIGKDD Conference on Knowledge Discovery and Data Mining},
  pages={5038--5049},
  year={2024}
}

@article{chen2021decision,
  title={Decision transformer: Reinforcement learning via sequence modeling},
  author={Chen, Lili and Lu, Kevin and Rajeswaran, Aravind and Lee, Kimin and Grover, Aditya and Laskin, Misha and Abbeel, Pieter and Srinivas, Aravind and Mordatch, Igor},
  journal={Advances in neural information processing systems},
  volume={34},
  pages={15084--15097},
  year={2021}
}

@inproceedings{liu2023constrained,
  title={Constrained decision transformer for offline safe reinforcement learning},
  author={Liu, Zuxin and Guo, Zijian and Yao, Yihang and Cen, Zhepeng and Yu, Wenhao and Zhang, Tingnan and Zhao, Ding},
  booktitle={International conference on machine learning},
  pages={21611--21630},
  year={2023},
  organization={PMLR}
}

@inproceedings{li2025gas,
  title={GAS: Generative Auto-bidding with Post-training Search},
  author={Li, Yewen and Mao, Shuai and Gao, Jingtong and Jiang, Nan and Xu, Yunjian and Cai, Qingpeng and Pan, Fei and Jiang, Peng and An, Bo},
  booktitle={Companion Proceedings of the ACM on Web Conference 2025},
  pages={315--324},
  year={2025}
}

@inproceedings{gao2025generative,
  title={Generative auto-bidding with value-guided explorations},
  author={Gao, Jingtong and Li, Yewen and Mao, Shuai and Jiang, Peng and Jiang, Nan and Wang, Yejing and Cai, Qingpeng and Pan, Fei and Jiang, Peng and Gai, Kun and others},
  booktitle={Proceedings of the 48th International ACM SIGIR Conference on Research and Development in Information Retrieval},
  pages={244--254},
  year={2025}
}

@inproceedings{he2021unified,
  title={A unified solution to constrained bidding in online display advertising},
  author={He, Yue and Chen, Xiujun and Wu, Di and Pan, Junwei and Tan, Qing and Yu, Chuan and Xu, Jian and Zhu, Xiaoqiang},
  booktitle={Proceedings of the 27th ACM SIGKDD Conference on Knowledge Discovery \& Data Mining},
  pages={2993--3001},
  year={2021}
}

@inproceedings{deng2021towards,
  title={Towards efficient auctions in an auto-bidding world},
  author={Deng, Yuan and Mao, Jieming and Mirrokni, Vahab and Zuo, Song},
  booktitle={Proceedings of the Web Conference 2021},
  pages={3965--3973},
  year={2021}
}

@article{aggarwal2024auto,
  title={Auto-bidding and auctions in online advertising: A survey},
  author={Aggarwal, Gagan and Badanidiyuru, Ashwinkumar and Balseiro, Santiago R and Bhawalkar, Kshipra and Deng, Yuan and Feng, Zhe and Goel, Gagan and Liaw, Christopher and Lu, Haihao and Mahdian, Mohammad and others},
  journal={ACM SIGecom Exchanges},
  volume={22},
  number={1},
  pages={159--183},
  year={2024},
  publisher={ACM New York, NY, USA}
}

@inproceedings{chen2011real,
  title={Real-time bidding algorithms for performance-based display ad allocation},
  author={Chen, Ye and Berkhin, Pavel and Anderson, Bo and Devanur, Nikhil R},
  booktitle={Proceedings of the 17th ACM SIGKDD international conference on Knowledge discovery and data mining},
  pages={1307--1315},
  year={2011}
}

@article{yu2017online,
  title={Online convex optimization with stochastic constraints},
  author={Yu, Hao and Neely, Michael and Wei, Xiaohan},
  journal={Advances in Neural Information Processing Systems},
  volume={30},
  year={2017}
}

@inproceedings{cai2017real,
  title={Real-time bidding by reinforcement learning in display advertising},
  author={Cai, Han and Ren, Kan and Zhang, Weinan and Malialis, Kleanthis and Wang, Jun and Yu, Yong and Guo, Defeng},
  booktitle={Proceedings of the tenth ACM international conference on web search and data mining},
  pages={661--670},
  year={2017}
}

@inproceedings{wen2022cooperative,
  title={A cooperative-competitive multi-agent framework for auto-bidding in online advertising},
  author={Wen, Chao and Xu, Miao and Zhang, Zhilin and Zheng, Zhenzhe and Wang, Yuhui and Liu, Xiangyu and Rong, Yu and Xie, Dong and Tan, Xiaoyang and Yu, Chuan and others},
  booktitle={Proceedings of the Fifteenth ACM International Conference on Web Search and Data Mining},
  pages={1129--1139},
  year={2022}
}

@article{mou2022sustainable,
  title={Sustainable online reinforcement learning for auto-bidding},
  author={Mou, Zhiyu and Huo, Yusen and Bai, Rongquan and Xie, Mingzhou and Yu, Chuan and Xu, Jian and Zheng, Bo},
  journal={Advances in Neural Information Processing Systems},
  volume={35},
  pages={2651--2663},
  year={2022}
}

@inproceedings{peng2025expert,
  title={Expert-Guided Diffusion Planner for Auto-Bidding},
  author={Peng, Yunshan and Shu, Wenzheng and Sun, Jiahao and Zeng, Yanxiang and Pang, Jinan and Bai, Wentao and Bai, Yunke and Liu, Xialong and Jiang, Peng},
  booktitle={Proceedings of the 34th ACM International Conference on Information and Knowledge Management},
  pages={5963--5970},
  year={2025}
}

@inproceedings{li2025ebaret,
  title={EBaReT: Expert-guided Bag Reward Transformer for Auto Bidding},
  author={Li, Kaiyuan and Wang, Pengyu and Peng, Yunshan and Yuan, Pengjia and Zeng, Yanxiang and Xiang, Rui and Cheng, Yanhua and Liu, Xialong and Jiang, Peng},
  booktitle={Companion Proceedings of the ACM on Web Conference 2025},
  pages={1104--1108},
  year={2025}
}

@inproceedings{zhao2018deep,
  title={Deep reinforcement learning for sponsored search real-time bidding},
  author={Zhao, Jun and Qiu, Guang and Guan, Ziyu and Zhao, Wei and He, Xiaofei},
  booktitle={Proceedings of the 24th ACM SIGKDD international conference on knowledge discovery \& data mining},
  pages={1021--1030},
  year={2018}
}

@article{balseiro2021robust,
  title={Robust auction design in the auto-bidding world},
  author={Balseiro, Santiago and Deng, Yuan and Mao, Jieming and Mirrokni, Vahab and Zuo, Song},
  journal={Advances in Neural Information Processing Systems},
  volume={34},
  pages={17777--17788},
  year={2021}
}

@inproceedings{ou2023deep,
  title={Deep landscape forecasting in multi-slot real-time bidding},
  author={Ou, Weitong and Chen, Bo and Yang, Yingxuan and Dai, Xinyi and Liu, Weiwen and Zhang, Weinan and Tang, Ruiming and Yu, Yong},
  booktitle={Proceedings of the 29th ACM SIGKDD Conference on Knowledge Discovery and Data Mining},
  pages={4685--4695},
  year={2023}
}

@article{borissov2010automated,
  title={Automated bidding in computational markets: an application in market-based allocation of computing services},
  author={Borissov, Nikolay and Neumann, Dirk and Weinhardt, Christof},
  journal={Autonomous Agents and Multi-Agent Systems},
  volume={21},
  number={2},
  pages={115--142},
  year={2010},
  publisher={Springer}
}

@inproceedings{nazerzadeh2008dynamic,
  title={Dynamic cost-per-action mechanisms and applications to online advertising},
  author={Nazerzadeh, Hamid and Saberi, Amin and Vohra, Rakesh},
  booktitle={Proceedings of the 17th international conference on World Wide Web},
  pages={179--188},
  year={2008}
}

@article{hu2016incentive,
  title={Incentive problems in performance-based online advertising pricing: Cost per click vs. cost per action},
  author={Hu, Yu and Shin, Jiwoong and Tang, Zhulei},
  journal={Management Science},
  volume={62},
  number={7},
  pages={2022--2038},
  year={2016},
  publisher={INFORMS}
}

@inproceedings{deng2023multi,
  title={Multi-channel autobidding with budget and roi constraints},
  author={Deng, Yuan and Golrezaei, Negin and Jaillet, Patrick and Liang, Jason Cheuk Nam and Mirrokni, Vahab},
  booktitle={International Conference on Machine Learning},
  pages={7617--7644},
  year={2023},
  organization={PMLR}
}

@inproceedings{korenkevych2024offline,
  title={Offline reinforcement learning for optimizing production bidding policies},
  author={Korenkevych, Dmytro and Cheng, Frank and Balakir, Artsiom and Nikulkov, Alex and Gao, Lingnan and Cen, Zhihao and Xu, Zuobing and Zhu, Zheqing},
  booktitle={Proceedings of the 30th ACM SIGKDD Conference on Knowledge Discovery and Data Mining},
  pages={5251--5259},
  year={2024}
}

@inproceedings{haarnoja2018soft,
  title={Soft actor-critic: Off-policy maximum entropy deep reinforcement learning with a stochastic actor},
  author={Haarnoja, Tuomas and Zhou, Aurick and Abbeel, Pieter and Levine, Sergey},
  booktitle={International conference on machine learning},
  pages={1861--1870},
  year={2018},
  organization={Pmlr}
}

@inproceedings{zheng2022online,
  title={Online decision transformer},
  author={Zheng, Qinqing and Zhang, Amy and Grover, Aditya},
  booktitle={international conference on machine learning},
  pages={27042--27059},
  year={2022},
  organization={PMLR}
}

@article{wu2026constraint,
  title={Constraint-Aware Generative Auto-bidding via Pareto-Prioritized Regret Optimization},
  author={Wu, Binglin and Zhang, Yingyi and Li, Xianneng and Deng, Ruyue and Yue, Chuan and Zhang, Weiru and Zeng, Xiaoyi},
  journal={arXiv preprint arXiv:2602.08261},
  year={2026}
}

@inproceedings{zhu2017optimized,
  title={Optimized cost per click in taobao display advertising},
  author={Zhu, Han and Jin, Junqi and Tan, Chang and Pan, Fei and Zeng, Yifan and Li, Han and Gai, Kun},
  booktitle={Proceedings of the 23rd ACM SIGKDD international conference on knowledge discovery and data mining},
  pages={2191--2200},
  year={2017}
}

@article{janner2021offline,
  title={Offline reinforcement learning as one big sequence modeling problem},
  author={Janner, Michael and Li, Qiyang and Levine, Sergey},
  journal={Advances in neural information processing systems},
  volume={34},
  pages={1273--1286},
  year={2021}
}

@inproceedings{wu2023masked,
  title={Masked trajectory models for prediction, representation, and control},
  author={Wu, Philipp and Majumdar, Arjun and Stone, Kevin and Lin, Yixin and Mordatch, Igor and Abbeel, Pieter and Rajeswaran, Aravind},
  booktitle={International Conference on Machine Learning},
  pages={37607--37623},
  year={2023},
  organization={PMLR}
}

@article{carroll2022uni,
  title={Uni [mask]: Unified inference in sequential decision problems},
  author={Carroll, Micah and Paradise, Orr and Lin, Jessy and Georgescu, Raluca and Sun, Mingfei and Bignell, David and Milani, Stephanie and Hofmann, Katja and Hausknecht, Matthew and Dragan, Anca and others},
  journal={Advances in neural information processing systems},
  volume={35},
  pages={35365--35378},
  year={2022}
}

@article{liu2022masked,
  title={Masked autoencoding for scalable and generalizable decision making},
  author={Liu, Fangchen and Liu, Hao and Grover, Aditya and Abbeel, Pieter},
  journal={Advances in Neural Information Processing Systems},
  volume={35},
  pages={12608--12618},
  year={2022}
}

@inproceedings{janner2022planning,
  title={Planning with Diffusion for Flexible Behavior Synthesis},
  author={Janner, Michael and Du, Yilun and Tenenbaum, Joshua and Levine, Sergey},
  booktitle={International Conference on Machine Learning},
  pages={9902--9915},
  year={2022},
  organization={PMLR}
}

@article{ajay2022conditional,
  title={Is conditional generative modeling all you need for decision-making?},
  author={Ajay, Anurag and Du, Yilun and Gupta, Abhi and Tenenbaum, Joshua and Jaakkola, Tommi and Agrawal, Pulkit},
  journal={arXiv preprint arXiv:2211.15657},
  year={2022}
}

@inproceedings{peebles2023scalable,
  title={Scalable diffusion models with transformers},
  author={Peebles, William and Xie, Saining},
  booktitle={Proceedings of the IEEE/CVF international conference on computer vision},
  pages={4195--4205},
  year={2023}
}

@article{tang2024learning,
  title={Learning versatile skills with curriculum masking},
  author={Tang, Yao and Xie, Zhihui and Lin, Zichuan and Ye, Deheng and Li, Shuai},
  journal={Advances in Neural Information Processing Systems},
  volume={37},
  pages={65562--65582},
  year={2024}
}

@article{vaswani2017attention,
  title={Attention is all you need},
  author={Vaswani, Ashish and Shazeer, Noam and Parmar, Niki and Uszkoreit, Jakob and Jones, Llion and Gomez, Aidan N and Kaiser, {\L}ukasz and Polosukhin, Illia},
  journal={Advances in neural information processing systems},
  volume={30},
  year={2017}
}

@inproceedings{perez2018film,
  title={Film: Visual reasoning with a general conditioning layer},
  author={Perez, Ethan and Strub, Florian and De Vries, Harm and Dumoulin, Vincent and Courville, Aaron},
  booktitle={Proceedings of the AAAI conference on artificial intelligence},
  volume={32},
  number={1},
  year={2018}
}

@inproceedings{he2022masked,
  title={Masked autoencoders are scalable vision learners},
  author={He, Kaiming and Chen, Xinlei and Xie, Saining and Li, Yanghao and Doll{\'a}r, Piotr and Girshick, Ross},
  booktitle={Proceedings of the IEEE/CVF conference on computer vision and pattern recognition},
  pages={16000--16009},
  year={2022}
}

@article{fujimoto2021minimalist,
  title={A minimalist approach to offline reinforcement learning},
  author={Fujimoto, Scott and Gu, Shixiang Shane},
  journal={Advances in neural information processing systems},
  volume={34},
  pages={20132--20145},
  year={2021}
}

@article{yao2011dynamic,
  title={A dynamic model of sponsored search advertising},
  author={Yao, Song and Mela, Carl F},
  journal={Marketing Science},
  volume={30},
  number={3},
  pages={447--468},
  year={2011},
  publisher={INFORMS}
}

@article{wilbur2009click,
  title={Click fraud},
  author={Wilbur, Kenneth C and Zhu, Yi},
  journal={Marketing Science},
  volume={28},
  number={2},
  pages={293--308},
  year={2009},
  publisher={INFORMS}
}

@article{dekimpe1995persistence,
  title={The persistence of marketing effects on sales},
  author={Dekimpe, Marnik G and Hanssens, Dominique M},
  journal={Marketing science},
  volume={14},
  number={1},
  pages={1--21},
  year={1995},
  publisher={INFORMS}
}

@article{sethuraman2011well,
  title={How well does advertising work? Generalizations from meta-analysis of brand advertising elasticities},
  author={Sethuraman, Raj and Tellis, Gerard J and Briesch, Richard A},
  journal={Journal of marketing research},
  volume={48},
  number={3},
  pages={457--471},
  year={2011},
  publisher={SAGE Publications Sage CA: Los Angeles, CA}
}

%%
%% If your work has an appendix, this is the place to put it.
\appendix
% Appendix

\section{SWAG-Bid Pipeline Algorithm}
\label{app:algorithm}

The complete pipeline of SWAG-Bid is summarized in Algorithm~\ref{alg:swag}, covering episode-level planning, dual-channel target transfer, and intra-episode execution. Since the planner runs only once per episode with four batched forward passes, its overhead is negligible for real-time bidding.

\renewcommand{\algorithmicrequire}{\textbf{Input:}}
\renewcommand{\algorithmicensure}{\textbf{Output:}}

\begin{algorithm}[htbp]
\caption{SWAG-Bid Pipeline}
\label{alg:swag}
\begin{algorithmic}[1]
\REQUIRE Trained planner $\pi_{\mathrm{plan}}$, trained controller $\pi_{\mathrm{ctrl}}$ with PSG-AdaLN, campaign duration $D$, per-episode budget $B_d$, constraint target $\rho_{\text{tgt}}$, window size $W$, number of samples $N$.
\ENSURE Bidding parameters $\{\lambda_{d,t}\}$ for all episodes and steps.

\FOR{episode $d = 1, \ldots, D$}
    \STATE Collect previous episode statistics; update history.
    \STATE Construct episode state $\bar{\mathbf{s}}_d$.
    \STATE // \textit{Episode-Level Planning (MWMS)}
    \FOR{stage $= 1, 2, 3, 4$}
        \STATE Apply stage-specific mask; forward $\pi_{\mathrm{plan}}$ to fill masked modality for $N$ samples in parallel.
    \ENDFOR
    \STATE Score each trajectory via Eq.~\eqref{eq:total_score}; select best: $\bar{a}_d^*$, $\bar{\rho}_d^*$.
    \STATE // \textit{Dual-Channel Target Transfer}
    \STATE $R_1 \leftarrow B_d / \bar{\rho}_d^*$; \quad $C_1 \leftarrow B_d$.
    \STATE Encode guidance: $\mathbf{g} \leftarrow \text{MLP}_{\text{guide}}(\bar{a}_d^*)$.
    \STATE // \textit{Intra-Episode Execution}
    \FOR{step $t = 1, \ldots, T$}
        \STATE Compute per-step gate $g_t$ via Eq.~\eqref{eq:gate}; interpolate $\tilde{\mathbf{g}}_t$ via Eq.~\eqref{eq:gate_interp}.
        \STATE $\lambda_{d,t} \leftarrow \pi_{\mathrm{ctrl}}(R_t, C_t, s_t, a_{<t};\; \tilde{\mathbf{g}}_t)$.
        \STATE Execute $\lambda_{d,t}$; observe $r_t, c_t$.
        \STATE $R_{t+1} \leftarrow R_t - r_t$; \quad $C_{t+1} \leftarrow C_t - c_t$.
    \ENDFOR
\ENDFOR
\end{algorithmic}
\end{algorithm}

\section{Implementation Details}
\label{app:impl}

All experiments are conducted on a compute cluster equipped with NVIDIA L20 GPUs. The model is trained using the AdamW optimizer. To ensure a fair comparison, all Transformer-based baselines share the same backbone capacity and are trained on the identical offline dataset. In particular, the strongest generative baselines GAS and PRO-Bid are trained with the same penalty function form $\phi(\rho, \rho_{\text{tgt}})$ as SWAG-Bid, applied to their per-episode efficiency ratios. We adopt the default hyperparameters from their original papers and tune them for optimal performance. All results are averaged over 10 independent runs with different random seeds.

The episode-level planner uses an Encoder-Decoder Transformer with $n_{\text{embd}} = 512$, $n_{\text{head}} = 8$, 2 encoder layers and 1 decoder layer, with a learning rate of $1 \times 10^{-4}$ and trajectory window length $L = 21$. The step-level controller uses a causal Transformer with $n_{\text{embd}} = 512$, 8 layers, $n_{\text{head}} = 16$, and context window $K = 20$, with a learning rate of $1 \times 10^{-5}$. PSG-AdaLN uses guidance dropout $p_{\text{drop}} = 0.2$, a maximum multiplicative noise scale of $0.3$, and guidance hidden size $H = 128$. For MWMS, the number of candidate samples is $N = 512$ and the confidence decay coefficient is $\kappa = 3.0$.

\section{Comparison of Guidance Injection Methods}
\label{app:conditioning}

To justify the design choice of PSG-AdaLN, we compare it against alternative guidance injection methods at 100\% budget: Concat (concatenate action target to input tokens), FiLM~\cite{perez2018film}, AdaLN-Zero~\cite{peebles2023scalable}, and Cross-Attention~\cite{vaswani2017attention}. Table~\ref{tab:conditioning} presents the results.

Concat performs the worst, indicating that simply appending the guidance to the input sequence does not effectively incorporate it into the model decision process. Cross-Attention and FiLM achieve moderate improvements, but they lack an explicit mechanism to modulate guidance reliance according to the intra-episode state. AdaLN-Zero, which applies static AdaLN modulation, is the strongest alternative, suggesting that modulation-based injection is more effective than concatenation or attention-based approaches. PSG-AdaLN further improves over AdaLN-Zero by a large margin, particularly in SW-ER, indicating that the per-step adaptive gate is important for the controller to dynamically adjust its reliance on guidance under varying market conditions and remaining budgets.

\begin{table}[htbp]
\centering
\caption{Comparison of guidance injection methods at 100\% budget.}
\vspace{-5pt}
\label{tab:conditioning}
\renewcommand{\arraystretch}{1.15}
\small
\begin{tabular}{lcc}
\toprule
Method & SW-Score $\uparrow$ & SW-ER $\downarrow$ \\
\midrule
Concat             & 217.9 & 0.321 \\
Cross-Attention~\cite{vaswani2017attention} & 222.5 & 0.289 \\
FiLM~\cite{perez2018film} & 228.6 & 0.294 \\
AdaLN-Zero~\cite{peebles2023scalable} & 230.6 & 0.262 \\
\textbf{PSG-AdaLN} & \textbf{240.7} & \textbf{0.208} \\
\bottomrule
\end{tabular}
\vspace{-5pt}
\end{table}

\section{Quantitative Robustness to Degraded Guidance}
\label{app:noise_robustness}

To complement the qualitative gate analysis in Sec.~\ref{sec:rq5}, we quantify end-to-end performance under corrupted guidance at 100\% budget: multiplicative noise of varying scales is injected into the action target at inference time. Table~\ref{tab:noise_robustness} presents the results.

Performance degrades gracefully as noise increases. Under 10\% noise, SW-Score drops by less than 1\% with a marginal rise in SW-ER. Even under 30\% noise, SWAG-Bid still attains a higher SW-Score than the strongest baseline operating with clean inputs (PRO-Bid, 227.1 in Table~\ref{tab:main_table}). This graceful degradation is consistent with the adaptive down-weighting behavior observed in Figure~\ref{fig:gate_heatmap}.

\begin{table}[htbp]
\centering
\caption{Performance under noisy guidance at 100\% budget.}
\vspace{-5pt}
\label{tab:noise_robustness}
\renewcommand{\arraystretch}{1.15}
\small
\begin{tabular}{lcc}
\toprule
Noise scale & SW-Score $\uparrow$ & SW-ER $\downarrow$ \\
\midrule
0\% (clean) & 240.7 & 0.208 \\
10\%        & 238.5 & 0.214 \\
30\%        & 234.1 & 0.238 \\
\bottomrule
\end{tabular}
\vspace{-5pt}
\end{table}

\section{Deployment Details of Online A/B Test}
\label{app:deploy}

To ensure the bidding policy adapts to the evolving e-commerce environment, both the planner and controller adopt a rolling window training strategy with separate data horizons: the planner is trained on logs from the most recent 90 consecutive days to capture long-term market trends, while the controller is trained on logs from the most recent 30 consecutive days to stay responsive to short-term dynamics. The planner performs inference once per day at the episode level, producing the bidding-intensity guidance for the current episode. To address the inherent long-tail distribution of real-world business data, both models apply logarithmic normalization of varying scales to different categories of features, ensuring numerical stability and better capture of high-variance samples.

To adapt SWAG-Bid to the online ROAS setting, where the objective is to maximize GMV while keeping the window-level cost-to-GMV ratio within the target, the RTG is redefined as the cumulative expected GMV and the CTG as the cumulative cost, while the PSG-AdaLN channel continues to carry planner bidding-intensity guidance. During online inference, the planner first runs to produce the episode-level guidance, which is then used to set the initial RTG for the controller (derived from the ROAS constraint target and planner efficiency prediction) and to provide the action guidance via PSG-AdaLN. The initial CTG is set to the per-episode budget. As the campaign progresses, these tokens are dynamically updated based on realized feedback to serve as real-time pacing signals. To satisfy the strict low-latency requirements of real-time bidding systems, we use the deterministic mean of the output action distribution as the final bidding parameter $\lambda_t$, avoiding the computational overhead of sampling.

% \section*{GenAI Usage Disclosure}
% During the preparation of this manuscript, the authors utilized Gemini 3.0 Pro exclusively for the purposes of improving readability, grammar, and spelling. No generative AI tools were used to formulate the research ideas, design the methodology, conduct the experiments, or generate the core scientific claims.

% The authors have meticulously reviewed and edited all text refined by the AI tool, and assume full and sole responsibility for the final content and integrity of this work.

% Generative AI tools were used only for spelling and grammar correction purposes. No content was generated or modified beyond these basic editing tasks.

%%
%% The acknowledgments section is defined using the "acks" environment
%% (and NOT an unnumbered section). This ensures the proper
%% identification of the section in the article metadata, and the
%% consistent spelling of the heading.
% \begin{acks}
% This work was supported by Alibaba Group through the Alibaba Innovative Research Program, and by the National Natural Science Foundation of China (NSFC) under Grants 72071029 and 72231010.
% \end{acks}

\end{document}